%% file: main.tex
\documentclass[10pt,twocolumn,letterpaper]{article}

\usepackage{cvpr} 
\usepackage[accsupp]{axessibility}
\input{preamble}

\definecolor{cvprblue}{rgb}{0.21,0.49,0.74}
\usepackage[pagebackref,breaklinks,colorlinks,allcolors=cvprblue]{hyperref}

\def\paperID{39} 
\def\confName{EarthVision}
\def\confYear{2026}

\title{How to Embed Matters:\\Evaluation of EO Embedding Design Choices}

\author{
Luis Gilch$^{1}$ \quad
Isabelle Wittmann$^{2}$ \quad
Maximilian Nitsche$^{1}$\\
Johannes Jakubik$^{2}$\quad
Arne Ewald$^{3}$\quad
Thomas Brunschwiler$^{2}$\\[6pt]
\small $^{1}$\,IBM Germany \quad $^{2}$\,IBM Research -- Europe \quad $^{3}$\,NORDAKADEMIE Germany\\
\small Corresponding author: \texttt{isabelle.wittmann1@ibm.com}
}

\begin{document}
\maketitle

\begin{abstract}
Earth observation (EO) missions produce petabytes of multispectral imagery, increasingly analyzed using large Geospatial Foundation Models (\mbox{GeoFMs}). Alongside end-to-end adaptation, workflows are making growing use of intermediate representations as task-agnostic embeddings, enabling models to compute representations once and reuse them across downstream tasks. Consequently, when \mbox{GeoFMs} act as feature extractors, decisions about how representations are obtained, aggregated, and combined affect downstream performance and pipeline scalability. Understanding these trade-offs is essential for scalable embedding-based EO workflows, where compact embeddings can replace raw data while remaining broadly useful.
We present a systematic analysis of embedding design in GeoFM‑based EO workflows. Leveraging NeuCo‑Bench, we study how backbone architecture, pretraining strategy, representation depth, spatial aggregation, and representation combinations influence EO task performance. We demonstrate the usability of GeoFM embeddings by aggregating them into fixed‑size representations more than 500× smaller than the raw input data. Across models, we find consistent trends: transformer backbones with mean pooling provide strong default embeddings, intermediate ResNet layers can outperform final layers, self‑supervised objectives exhibit task‑specific strengths, and combining embeddings from different objectives often improves robustness.
\end{abstract}

\section{Introduction}
\label{sec:intro}
Earth observation (EO) from space produces a continuously growing stream of high-resolution, multispectral imagery used for climate monitoring, disaster response, agriculture, and urban planning~\cite{guo_earth_2015}. Since the launch of Landsat-1 in 1972, satellite missions such as the European Copernicus Sentinels have expanded the availability of free, high-cadence imagery, driving widespread adoption in academia and industry~\cite{esa2025}. Today, EO archives exceed hundreds of petabytes and continue to grow rapidly, increasing the demand for scalable analysis methods~\cite{wilkinson_environmental_2024}.

In parallel, machine learning for EO has evolved from pixel-level classifiers and handcrafted features to convolutional and transformer-based architectures that learn high-dimensional feature representations~\cite{zhu2017deep, wang2022ssl}. This shift has led to Geospatial Foundation Models (\mbox{GeoFMs}), pretrained on large-scale EO corpora and adapted to downstream tasks through end-to-end fine-tuning~\cite{bommasani_opportunities_2021, xiao2025foundation}. While effective, such pipelines require repeated access to raw imagery and backbone models, incurring computational and storage costs.
To mitigate these limitations, \mbox{GeoFMs} can instead be used to extract generic embeddings that are reused across multiple downstream tasks, offering improved computational efficiency, lightweight decoders, decentralized deployment, and similarity search at scale. In this embedding-centric paradigm, the backbone serves as a fixed feature extractor whose representations are computed once and reused across tasks~\cite{klemmer2025earth}.
This reduces the need to repeatedly access and process raw imagery, enabling compact embedding stores that can replace raw data with effective compression ratios, depending on the downstream task and representation design. At EO scale, this supports efficient storage, I/O, and retrieval workflows, particularly when dense spatial or temporal representations are aggregated into compact fixed‑size embeddings. 

With the rapid release and adoption of new \mbox{GeoFMs}, embeddings derived from these models are becoming central artifacts in EO pipelines. In this setting, design decisions about where representations are extracted, how they are spatially or temporally aggregated, and which information is retained directly shape downstream performance, storage footprint, and scalability. Yet it remains unclear which representation characteristics matter most for different tasks, which aggregation strategies are most effective, and how \mbox{GeoFM} outputs can be transformed into compact, fixed‑size embeddings that remain broadly useful across applications. As a result, principled design of compact and effective EO embeddings remains an open challenge.
\begin{figure*}[t]
\centering
\begin{subfigure}{0.47\textwidth}
\centering
\includegraphics[width=\linewidth]{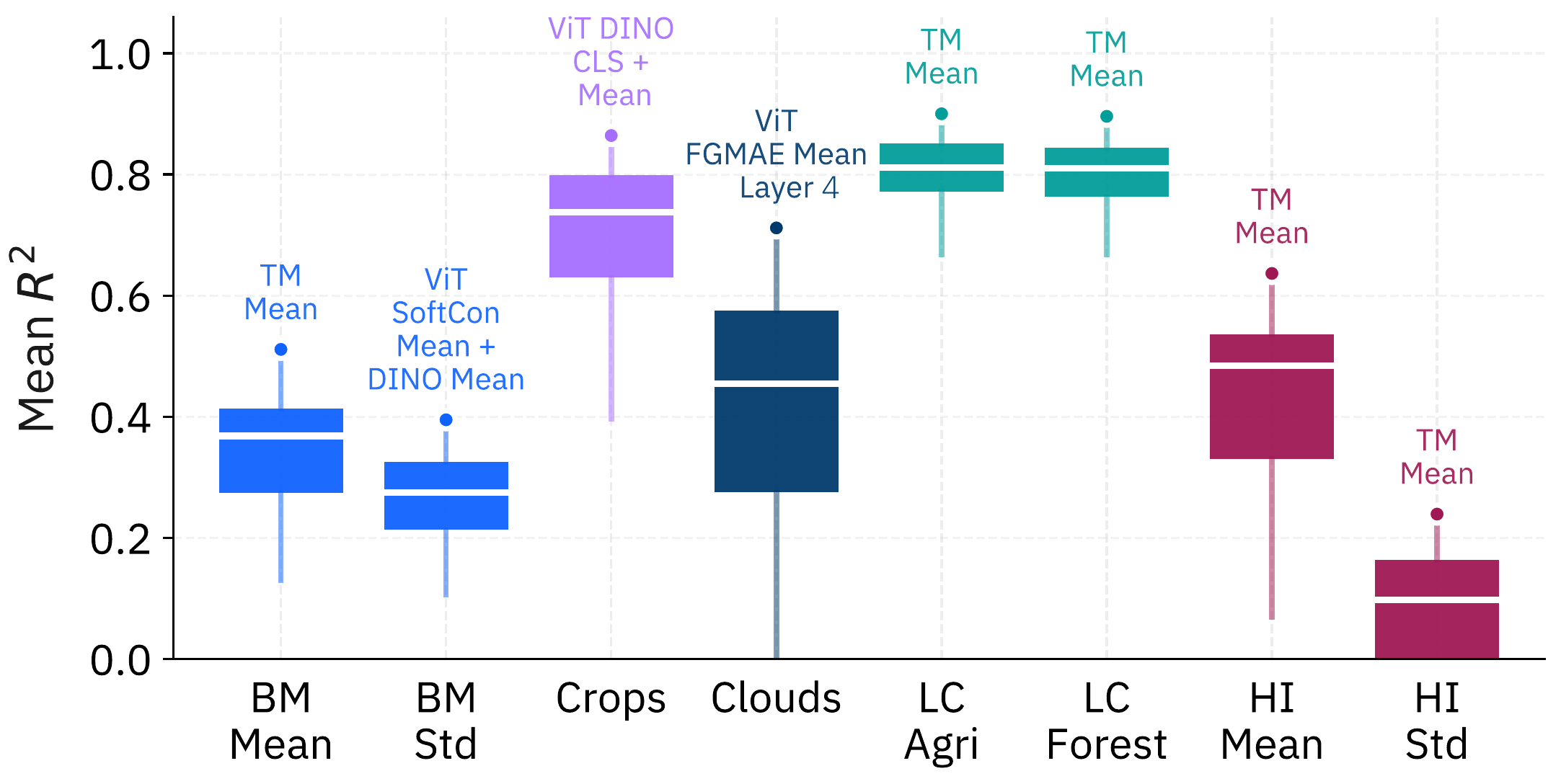}
\end{subfigure}\hfill
\begin{subfigure}{0.47\textwidth}
\centering
\includegraphics[width=\linewidth]{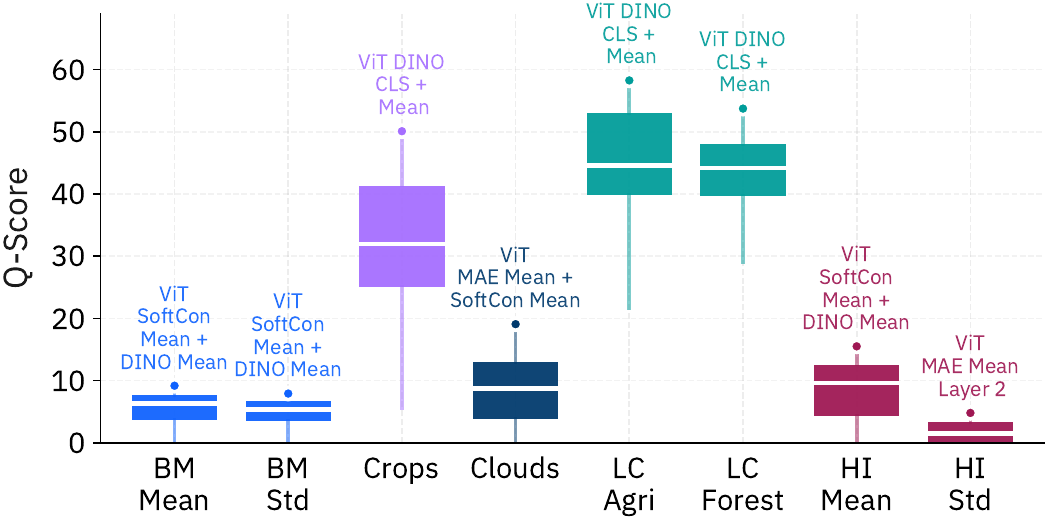}
\end{subfigure}
\caption{
\textbf{Per-task embedding performance across design choices.}
Distribution of regression performance across GeoFM backbones, self-supervised pretraining strategies, spatial aggregation methods, intermediate layers, and representation combinations. 
Performance is measured using mean $R^2$ (left), reflecting predictive accuracy, and the NeuCo Quality Score (right), which accounts for variability to reflect robustness. 
While most methods achieve similar peak accuracy for more saturated tasks, robustness varies, leading to clearer differentiation in the Quality Score; concatenated representations often rank among the most robust configurations in these cases. Boxplots summarize the distribution over all evaluated embedding variants; whiskers denote $1.5\times$ IQR and outliers are omitted. Markers indicate the single best-performing embedding configuration among all evaluated variants for the respective task.
}
\label{fig:task_spread}
\end{figure*}

\section{Related Work}
\label{sec:related_work}

\paragraph{Geospatial Foundation Models.}
Recent advances in large-scale self-supervised learning on multispectral EO data have led to the emergence of Geospatial Foundation Models (\mbox{GeoFMs})~\cite{xiao2025foundation, wang_ssl4eo-s12_2023, wilkinson_environmental_2024}. Early models use convolutional backbones such as ResNets~\cite{hoeser_object_2020, xiong_earthnets_2022}, which encode locality priors, while newer approaches adopt Vision Transformers (ViTs)~\cite{dosovitskiy_image_2020, wang_ssl4eo-s12_2023} to capture long‑range dependencies via self‑attention. Multimodal \mbox{GeoFMs} further extend these architectures to integrate multi-modal EO inputs~\cite{dofa, jakubik_terramind_2025}. %
Self‑supervised learning (SSL) underpins most \mbox{GeoFMs}, spanning contrastive methods~\cite{he_momentum_2020, caron_emerging_2021}, masked reconstruction approaches~\cite{he_masked_2022, wang_feature_2024}, and extensions tailored to multimodal or multi‑label EO data~\cite{wang_multi-label_2024, wang_decoupling_2024}. \mbox{GeoFMs} are typically evaluated via end-to-end fine-tuning with task-specific decoders.

\paragraph{Embedding-Centric Workflows in EO.}
EO research increasingly publishes and leverages precomputed vector representations (embeddings) that summarize scene content and enable efficient downstream use~\cite{klemmer2025earth}. These task‑agnostic embeddings capture general spatial, spectral, or spatiotemporal structure and support broad reuse. 
Examples include SatCLIP~\cite{klemmer_satclip_2025}, which derives global location embeddings via contrastive image-coordinate alignment; TESSERA~\cite{feng2025tessera}, which produces pixel-wise temporal embeddings from multi-temporal observations; and AlphaEarth Foundations~\cite{brown2025alphaearth}, which models continuous spatiotemporal embedding fields to summarize multi-sensor EO. 
These efforts advance an embedding-centric paradigm in which \mbox{GeoFMs} are trained to produce reusable representations as standalone data products. In parallel, many existing \mbox{GeoFMs} developed for end‑to‑end adaptation are increasingly used as off-the-shelf feature extractors~\cite{czerkawski_global_2024}, with the backbone acting as a frozen encoder. However, design choices around representation depth, aggregation, and fixed‑size construction remain largely heuristic.
In contrast, we systematically analyze embedding extraction from frozen \mbox{GeoFMs}, examining how architecture, layer selection, pooling, and pretraining objective influence downstream performance and robustness, providing guidance for embedding-centric GeoFM design and use.

\paragraph{Benchmarks for \mbox{GeoFMs} and Embeddings.} Benchmarking \mbox{GeoFMs} typically relies on diverse downstream tasks. Two established GeoFM benchmarks, PANGEA~\cite{marsocci_pangaea_2024} and GEO-BENCH~\cite{lacoste_geo-bench_2023, simumba2026geobench2}, span diverse geospatial domains and task types, including segmentation, classification, change detection, and regression. While differing in design choices, both assume access to the backbone during downstream adaptation allowing downstream models to reuse intermediate feature maps through multi-scale encoder features (e.g., in U-Net-style decoders), rather than constraining the representation to a single fixed-size embedding as in our setting.
In contrast, NeuCo-Bench~\cite{vinge2025neuco}, which we use in this work, is a model-agnostic framework for evaluating compact fixed-size embeddings. It comprises per-image regression and classification tasks spanning land cover, biomass, clouds, and heat-island prediction, and is explicitly designed to assess EO embeddings derived from spatiotemporal inputs. Section~\ref{sec:methodology} details how we use NeuCo‑Bench for our embedding design analysis.

\section{Methodology}
\label{sec:methodology}
We study embedding‑centric EO workflows in which pretrained \mbox{GeoFMs} serve as frozen feature extractors, evaluating their representations across diverse downstream tasks. Our analysis focuses on how embedding design choices, including encoder family, intermediate layer selection, spatial and temporal aggregation, and representation combinations, shape downstream regression performance and robustness. All embeddings are evaluated using the NeuCo‑Bench framework~\cite{vinge2025neuco}.

\subsection{Evaluation Protocol}

\paragraph{NeuCo-Bench Setup.}
NeuCo‑Bench is a framework for fixed‑size EO embeddings. Encoders are treated as black boxes: for each sample, we compute an embedding using a defined extraction method and evaluate it via linear probing on downstream tasks. In the original NeuCo-Bench challenge, embeddings were constrained to a fixed dimensionality (e.g., 1024) to reflect compression limits. Here, we instead focus on representation design and evaluate embeddings at their native dimensionality. Even without an explicit dimensionality constraint, these embeddings remain highly compact, corresponding to effective compression ratios of roughly 500× to over 2000× relative to the raw Sentinel-2 L1C patch data. When embedding dimensionality differs in controlled comparisons (e.g., intermediate‑layer analyses), we introduce resized baselines to ensure that performance differences are not attributable to embedding size alone.

\paragraph{Cross-Validation and Metrics.}
For each embedding method $p$ and task $t$, we perform $K=50$ repeated random train-test splits. On each split $k$, a linear regressor is trained on the training subset and evaluated on the held-out data, yielding a test-set $R^2$ score, $s_k$.
Performance is summarized by the mean across splits,
\[
\bar{R}^2_{t,p} = \frac{1}{K} \sum_{k=1}^{K} s_k,
\]
measuring average predictive accuracy.
Additionally, we report the NeuCo-Bench Quality Score~\cite{vinge2025neuco},
\[
Q_{t,p} = 100\,\epsilon\,\frac{\bar{R}^2_{t,p}}{\sigma_{t,p} + \epsilon},
\]
where $\sigma_{t,p}$ denotes the standard deviation of $s_k$ across splits and $\epsilon = 0.02$ ensures numerical stability.
While $\bar{R}^2_{t,p}$ reflects predictive strength, $Q_{t,p}$ captures both accuracy and variability. An embedding method with slightly lower mean $R^2$ but substantially lower variance can therefore achieve a higher Quality Score, indicating more stable generalization across data splits.
All experiments use identical linear probing settings, with batch size 64, 20 epochs, learning rate $10^{-3}$, AdamW optimizer, MSE loss, and 50 repeated splits.
\paragraph{Benchmark Dataset.}
The public NeuCo-Bench tasks are based on the SSL4EO-S12-downstream dataset~\cite{ssl4eo_downstream}, which follows the data structure of SSL4EO-S12~\cite{wang_ssl4eo-s12_2023, blumenstiel2026ssl4eos12v11}. Each input sample consists of four seasonal timesteps with 27 available bands from Sentinel-1 and Sentinel-2 (L2A and L1C), forming a $264 \times 264$ patch at 10m resolution. Each spatiotemporal input cube is paired with a scalar target.
We evaluate embeddings on eight regression tasks:
\textit{Biomass Mean + Std} (GEDI‑derived above‑ground biomass statistics),
\textit{Crops} (corn and soybean coverage),
\textit{Clouds} (mean cloud cover),
\textit{Land-cover agricultural + forest} (area fractions), and
\textit{Heat Island Mean + Std} (urban surface temperature statistics). 
Although all tasks are formulated as aggregated regression targets, their underlying signal characteristics differ. We distinguish between semantic proportion tasks, continuous physical measurement tasks, and atmospheric state indicators.

Land-cover (Agriculture, Forest) and Crops represent aggregated semantic proportions derived from categorical land-cover maps, where the target reflects the fraction of a patch assigned to discrete classes. Biomass and Heat Island originate from continuous physical measurements
(LiDAR-based biomass estimates and thermal surface temperature) that are spatially aggregated to patch-level mean and standard deviation statistics. Clouds, while also modeled as a continuous regression target, is derived from per-acquisition cloud masks and reflects temporally varying atmospheric state rather than a direct surface biophysical property.
These differing semantic, structural, and temporal characteristics enable analysis of task-consistent embedding trends across different signal types.

\subsection{GeoFM Backbones and Pretraining Strategies}

We evaluate SSL4EO-pretrained ResNet-50~\cite{he2016resnet} and ViT-Small~\cite{dosovitskiy_image_2020} backbones~\cite{wang_ssl4eo-s12_2023} to maintain comparability in pretraining data. We include TerraMind (ViT-Small)~\cite{terramind_small_hf} as a reference, as its TerraMind-Base~\cite{jakubik_terramind_2025} variant achieved leading performance in the evaluations in the original NeuCo-Bench paper. SSL4EO-pretrained backbones are accessed via TorchGeo~\cite{stewart2025torchgeo}, while TerraMind weights are accessed via TerraTorch~\cite{gomes2025terratorch}. Embeddings are extracted with TerraTorch for consistent processing across models.
For the SSL4EO-pretrained backbones, we consider the self-supervised methods DINO~\cite{caron_emerging_2021}, MoCo~\cite{he_momentum_2020}, DECUR~\cite{wang_decoupling_2024}, SoftCon~\cite{wang_multi-label_2024}, MAE~\cite{he_masked_2022}, and FGMAE~\cite{wang_feature_2024}. Not all backbone-method combinations are evaluated due to limited availability in pretrained weights. To ensure comparability, all models are evaluated using Sentinel‑2 L1C inputs only, although TerraMind supports additional modalities.

\subsection{Temporal and Spatial Aggregation}

The evaluated backbones are non-temporal encoders; each of the four seasonal observations is therefore encoded independently.
Let $x^{(i)}$ denote the $i$-th seasonal input and $f(\cdot)$ the frozen encoder. We compute
\[
z^{(i)} = f(x^{(i)}),
\]
and aggregate the resulting latent vectors using mean pooling:
\[
z = \frac{1}{4} \sum_{i=1}^{4} z^{(i)}.
\]
This yields a single temporally aggregated embedding per sample. 

The considered GeoFM encoders produce dense spatial representations: ResNet outputs feature maps of shape $C \times H \times W$, while ViT models produce patch tokens of shape $N_{\text{patch}} \times D$. Image inputs are resized to $224 \times 224$ to match pretraining resolution. To obtain a 1D fixed‑size embedding per sample, we apply global mean pooling, across spatial dimensions for ResNet and across tokens for the ViT. We additionally evaluate max and min pooling for both architectures, and for ViTs also analyze CLS‑token embeddings. Embeddings are evaluated at their native dimensionality (2048 for ResNet‑50 and 384 for ViT‑Small).
\begin{figure*}[t]
  \centering
  \includegraphics[width=0.75\textwidth]{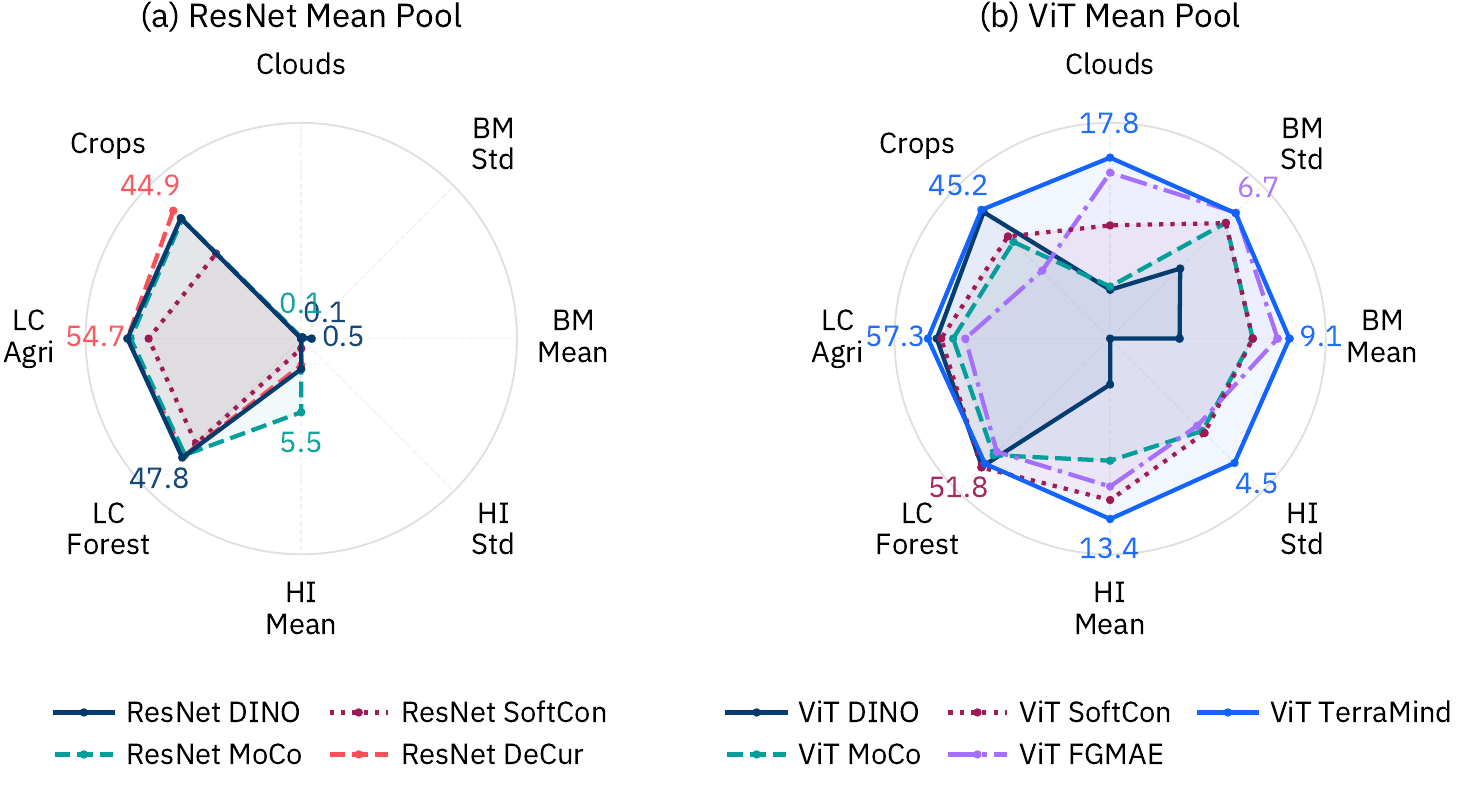}
  \caption{\textbf{Per-task Q-score comparison of ResNet-50 (left) and ViT-Small (right) FMs.} We use final-layer embeddings with mean pooling; negative scores are clipped to zero. ResNet models score high on semantic/land-cover tasks but show little performance elsewhere. ViT models are more consistent across tasks and achieve meaningful performance beyond land cover: TerraMind is the most consistent overall, DINO is strong on land cover but weaker on other tasks, and FGMAE excels on the cloud-cover and biomass tasks. Radial axis is centered at $0$, and the maximal radius is set globally with a fixed buffer. A corresponding task $R^2$ plot is provided in the supplementary.}
  \label{fig:radar_methods}
\end{figure*}
\subsection{Concatenation Experiments}
We evaluate whether combining representations improves downstream performance. Concatenation experiments are restricted to three ViT‑based SSL methods (DINO, MAE, and SoftCon). We test (i) concatenating the mean-pooled embedding and the CLS token embedding from the same ViT model, and
(ii) concatenating mean‑pooled embeddings from pairs of ViT models. The first setting assesses whether different token‑aggregation strategies within a model yield complementary information, while the second examines whether embeddings learned under different SSL objectives can be effectively combined. 

\subsection{Intermediate-Layer Analysis}
To assess representation quality across encoder depth, we evaluate intermediate features. For ViT-Small, we extract token representations after each of the 12 transformer blocks (constant embedding size). For ResNet-50, we extract feature maps after each residual stage (64--2048 channels). As before, spatial pooling is applied, and each embedding is assessed at its native dimensionality. Because CNN feature dimensionality increases with depth, we additionally include a resized final-layer baseline for ResNet models. This baseline is constructed by channel-wise averaging of the final-layer embedding to match the dimensionality of the corresponding intermediate representation.

\section{Experiments and Results}
\label{sec:expirments}
We evaluate embedding design choices for \mbox{GeoFMs} across backbone choice, pretraining strategy, spatial aggregation, and representation depth. We begin with global per-task trends before analyzing each design dimension.

\subsection{General Per-Task Trends}

We first examine performance across all evaluated embedding variants in Figure~\ref{fig:task_spread}, aggregating results over backbones, SSL objectives, pooling strategies, intermediate layers, and concatenation settings.

\textbf{Semantic land-cover targets.}
Land-cover Agriculture and Land-cover Forest achieve the highest median $R^2$ values (both above 0.80) with low variability, indicating that these signals are reliably captured across embeddings. Crops shows moderate spread. While peak $R^2$ values are partially saturated, the Q-score reveals meaningful differences in robustness not visible from mean accuracy alone.

\textbf{Continuous biophysical and atmospheric targets} are more challenging. Biomass and Heat Island Mean show lower median $R^2$, and Clouds exhibits strong sensitivity to representation design with wide variation across methods, indicating greater dependence on representation choice. Heat Island Std is the most difficult target overall.

Comparing $R^2$ and Q-score further separates accuracy from robustness. TerraMind mean embeddings often achieve the highest average $R^2$, whereas the strongest Q-scores frequently arise from concatenated representations, suggesting that combining complementary embeddings improves stability. Overall, task difficulty varies substantially, and robustness provides complementary insight beyond mean $R^2$ when comparing embedding strategies.

\subsection{Transformer vs. CNNs}
\label{main:backbone}
In the Q‑score radar plot (Figure~\ref{fig:radar_methods}), we compare ResNet‑50 and ViT‑Small backbones using mean‑pooled embeddings ($R^2$ radar plots and full tables are provided in the supplementary material). A consistent pattern emerges: ResNet models systematically underperform on geophysical and atmospheric targets (Biomass, Clouds, Heat Island), where ViT models achieve substantially stronger results.

\textbf{Continuous biophysical and atmospheric targets.} Across ResNet configurations, $R^2$ remains near zero or negative for Biomass, Clouds, and Heat Island Std. Even the strongest ResNet variant (DINO) reaches only $R^2 = 0.05$ on Biomass Mean and $R^2 = -0.20$ on Clouds. In contrast, ViT models achieve markedly higher scores with up to 0.50 on Biomass Mean and 0.69 on Clouds. This separation is mirrored in Q‑scores, indicating that the backbone gap persists when robustness across splits is considered.

\textbf{Semantic land‑cover targets.} For Crops, Land-cover Agriculture, and Land-cover Forest, ResNets remain competitive and in some cases slightly exceed ViT performance. This suggests that global semantic composition signals are less sensitive to backbone choice, whereas geophysical and atmospheric variables benefit from richer, long‑range spatial modeling. Across all configurations, TerraMind (ViT‑Small with multimodal pretraining on TerraMesh~\cite{blumenstiel_terramesh_2025}) achieves the strongest and most stable performance. 

These trends align with prior NeuCo‑Bench backbone comparisons performed under matched embedding dimensionality~\cite{vinge2025neuco}, where CNN embeddings were downsampled and ViT embeddings padded. Together with our results, this indicates that the observed ResNet-ViT gap cannot be attributed to embedding dimensionality or resizing artifacts.

\subsection{Task-Specific Effects of SSL Objectives}
\label{main:ssl}
Beyond backbone architecture, the self-supervised pretraining objective shapes task performance. For ViT-Small models pretrained on SSL4EO (Q-score radar plot in Figure~\ref{fig:radar_methods}; $R^2$ in the supplementary material), no task-agnostic ranking emerges. Instead, each objective shows a distinct performance profile.
DINO performs strongest on semantic land-cover proportion targets (Crops, LC Agriculture, LC Forest). In contrast, MAE and FGMAE achieve higher scores on Biomass and Clouds, suggesting that reconstruction-based objectives better capture continuous biophysical variation. SoftCon shows the most balanced behavior and achieves the highest average performance among the SSL4EO ViT models.

These trends are consistent across both $R^2$ and Q-score. While $R^2$ reflects accuracy, the Q-score magnifies robustness differences, and in our results leads to clearer separation between methods.
Objective-dependent differences are also more pronounced for transformer backbones. For ResNet models, variation across SSL objectives is weaker, as all methods struggle on continuous biophysical and atmospheric targets.
Overall, pretraining objective shapes task-specific representation profiles. Contrastive objectives (e.g., DINO) favor semantic composition signals, whereas reconstruction-based objectives (e.g., MAE/FGMAE) better capture continuous geophysical variation. Rather than yielding a single universally best method, SSL objectives produce complementary strengths, an observation we will revisit in the concatenation experiments.

\subsection{Impact of Spatial Pooling}
\label{main:pooling}
\begin{figure}[t]
  \centering
  \includegraphics[width=0.98\linewidth]{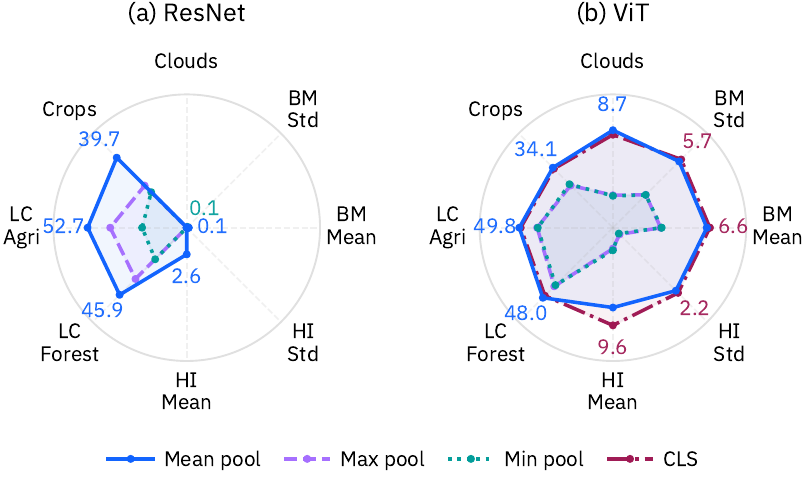}
    \caption{\textbf{Per-task Q-score comparison of spatial aggregation method for ResNet-50 (left) and ViT-Small (right).} We use final-layer embeddings with mean, min, or max pooling (or the CLS token for ViT) and average scores across models; negative scores are clipped to zero. For ResNet, mean pooling performs best across tasks, with max pooling outperforming min pooling. For ViT, mean pooling again performs best, with CLS comparable on most tasks, while min and max pooling are similar but weaker, especially on continuous biophysical targets. A corresponding task $R^2$ plot is provided in the supplementary material.}
  \label{fig:radar_pooling}
\end{figure}

Figure~\ref{fig:radar_pooling} compares spatial pooling strategies per backbone using Q‑score (with corresponding $R^2$ results in the supplementary material). We evaluate mean, min, and max pooling for both architectures, and additionally the CLS token for ViT.

\textbf{Across both backbones}, mean pooling consistently provides the most reliable performance. It achieves the highest average scores and remains strong on both semantic land‑cover and continuous environmental targets. In contrast, min and max pooling generally underperform, particularly on continuous biophysical targets such as Biomass, Clouds, and Heat Island, suggesting that extreme-value aggregation discards some spatial information that is critical.

\textbf{For ViT models}, CLS pooling remains competitive with mean pooling and delivers comparable overall performance. It slightly improves certain regression targets (e.g., Heat Island Mean, Biomass) while underperforming on others. A task‑specific exception appears for ViT‑DINO, where min/max pooling slightly improves Biomass Mean (0.37 vs.\ 0.28), but this pattern does not generalize.

Overall, mean pooling emerges as the most robust and broadly applicable strategy, with CLS pooling offering a competitive alternative for transformer‑based models. These trends hold consistently across both $R^2$ and Q‑score.

\begin{figure*}[t]
\centering
\includegraphics[width=0.92\textwidth]{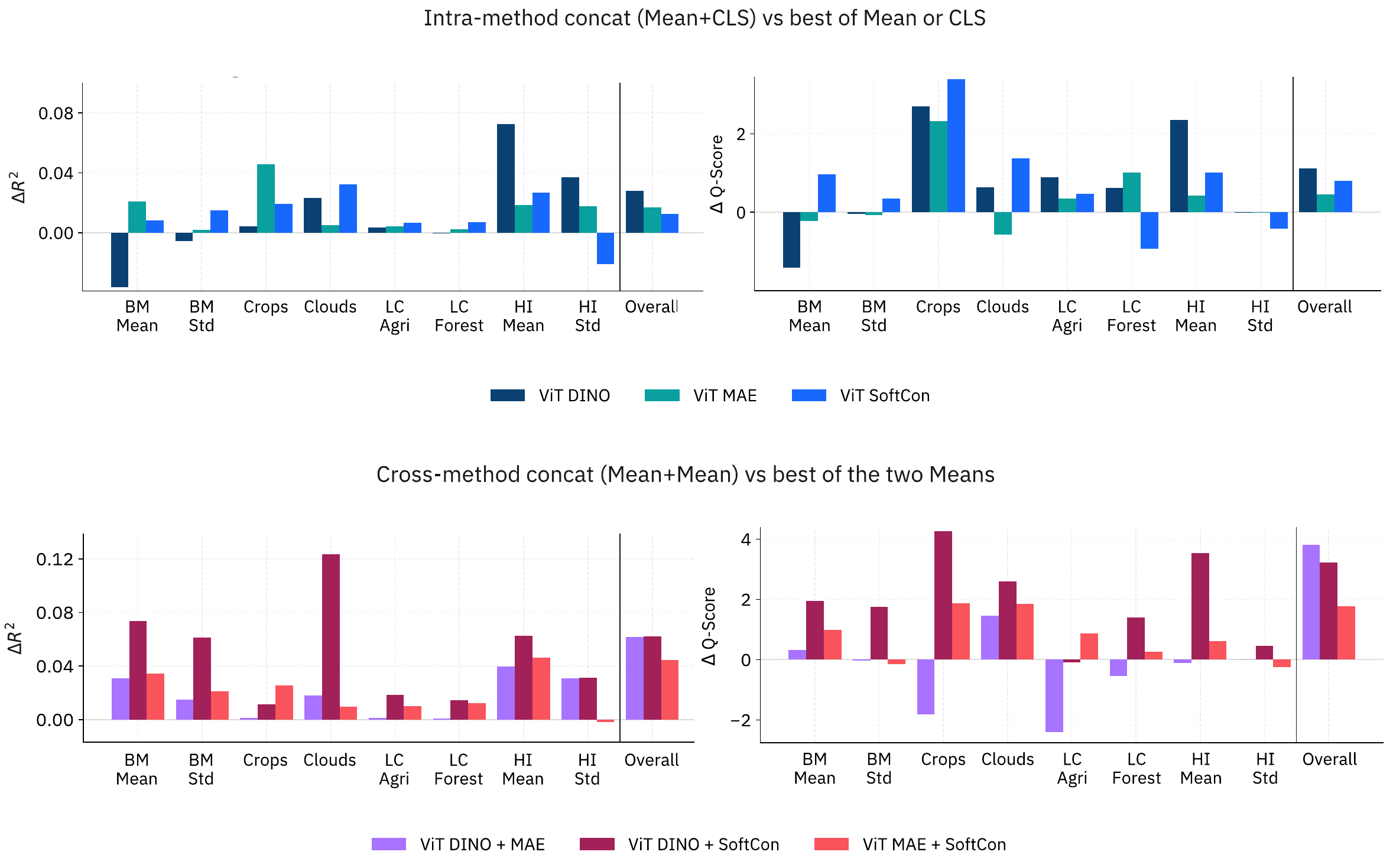}
\caption{\textbf{Per-task and overall $\Delta R^2$ (left) and $\Delta$ Q-score (right) for embedding concatenation.}
Top: \emph{Intra-method} concatenation (Mean + CLS within the same ViT-Small SSL4EO model).
Bottom: \emph{Inter-method} concatenation (Mean + Mean across different SSL objectives).
For each task, the baseline is the stronger individual embedding, and we report $\Delta = \text{score}_{\text{concat}} - \text{score}_{\text{baseline}}$ (zero indicates no change).
We additionally report the overall gain relative to the stronger overall baseline.
Intra-method (Mean+CLS) concatenation yields only modest improvements (typically $<0.04$ in $R^2$ and $<1$ Q-score point), indicating substantial redundancy between token aggregation strategies.
In contrast, inter-method (Mean+Mean) concatenation produces larger overall gains and consistent per-task improvements, reflecting complementary strengths across SSL objectives.
While per-task deltas remain moderate, overall gains demonstrate that diversity in pretraining objectives contributes more to complementarity than alternative token aggregation within a single model.}

\label{fig:concat_all}
\end{figure*}

\subsection{Complementarity Through Concatenation}
\label{main:concat}

We analyze embedding concatenation from two perspectives:
(i) per-task gains relative to the stronger task-specific baseline, and
(ii) overall gains relative to the strongest baseline across tasks.
These two measurements capture distinct effects.
When concatenating embeddings under the multi‑task NeuCo‑Bench setting, two effects are possible. First, if individual baselines excel on different tasks, concatenation can improve the overall score simply by retaining the stronger signal per task, even when per‑task gains remain small or negative. Second, if the representations encode genuinely complementary information, concatenation may also yield per‑task improvements beyond both baselines, resulting in gains at both levels.

\paragraph{Intra-method concatenation (Mean + CLS).}

Across DINO, MAE, and SoftCon, intra‑method concatenation yields only modest improvements (below $\sim$0.03 R$^2$ and below 1 Q‑score point overall), with consistently small per‑task gains (Figure~\ref{fig:concat_all}). This matches our earlier observation that Mean and CLS embeddings from the same ViT model show highly aligned task profiles and similar overall performance. Given this overlap, few complementary task scores can be exploited for overall improvements.

Our primary interest is whether CLS+Mean captures complementary signals that benefit individual tasks. The consistently small per-task improvements, however, suggest substantial redundancy between the two aggregation strategies. One notable exception is the comparatively large gain of DINO Mean+CLS on HeatIsland Mean.
While $R^2$ and Q-score trends largely align, some combinations (e.g., ViT SoftCon Mean+CLS) exhibit slightly stronger gains in Q-score (e.g., on the high-variance Cloud task). This indicates improved robustness, even when changes in mean predictive performance remain modest.

\paragraph{Cross-method concatenation (Mean + Mean).}

\begin{figure*}[t]
    \centering
    \includegraphics[width=0.76\textwidth]{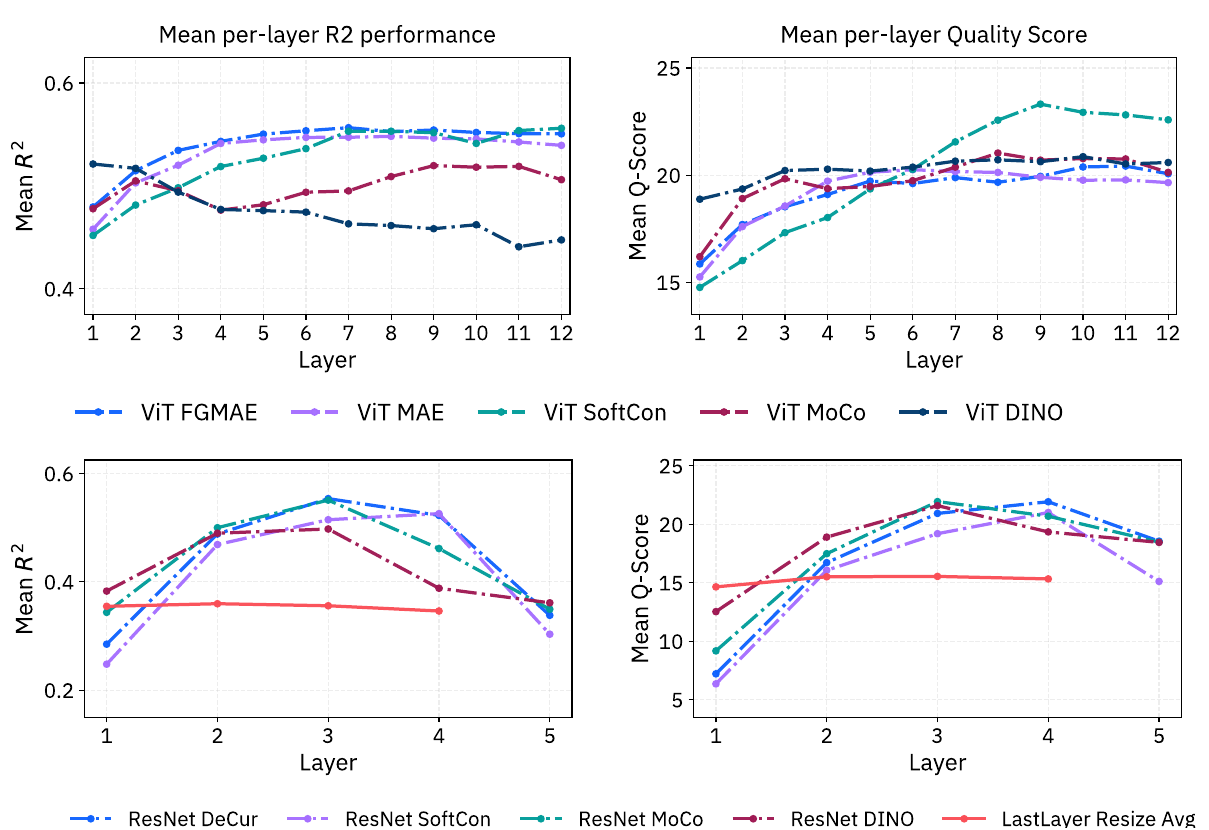}
    \caption{
    \textbf{Layer-wise task-averaged performance ($R^2$, left; Q-score, right).}
    Top: ViT-Small; bottom: ResNet-50 (SSL4EO). 
    Representations are extracted from each layer (12 transformer blocks; 5 ResNet stages with output dimensions 64, 256, 512, 1024, 2048); negative task values are clipped before averaging. 
    ViT performance increases in early layers and then saturates, whereas ResNet shows an inverted-U pattern, peaking at intermediate stages and degrading at the final layer. 
    A resized last-layer reference is included for ResNet.
    }
    \label{fig:layer_all}
\end{figure*}
Cross-method concatenation produces substantially larger overall improvements (approximately 0.04--0.07 R$^2$ and 1--4 Q‑score points; see Figure~\ref{fig:concat_all}). For context, individual ViT mean baselines vary by up to $\sim$0.11 in average R$^2$ ($\sim$0.19 including TerraMind), placing these gains on the same scale as differences between SSL objectives themselves.

These stronger overall gains reflect the greater diversity among ViT mean baselines compared to Mean vs.\ CLS within a model. Prior results show that DINO excels on semantic targets, SoftCon exhibits a balanced profile, and MAE is particularly strong on Clouds while weaker on Crops. Concatenation leverages these complementary strengths to increase overall performance.
This aggregation effect is especially clear for DINO + MAE: per‑task gains relative to the stronger baseline remain small, and in Q‑score occasionally negative on semantic tasks dominated by DINO, yet overall performance improves substantially. Here, concatenation rarely surpasses the strongest baseline per task but effectively combines strengths across tasks.

Beyond aggregation, certain combinations show genuine synergies. Most notably, DINO + SoftCon achieves visible per‑task improvements beyond both baselines on Clouds and consistent gains on Biomass and Heat Island Mean. These improvements are also reflected in Q‑score, indicating increased robustness.

\paragraph{} These results show that concatenation benefits arise primarily from differences in SSL pretraining objectives rather than from token aggregation strategies (Mean vs.\ CLS). When baseline task profiles are highly aligned, as with Mean and CLS from the same model, gains remain limited. When baselines exhibit complementary strengths (DINO, MAE, SoftCon), concatenation can consolidate these strengths to improve overall scores and, in some cases, deliver per‑task improvements beyond both baselines.

\subsection{Strength of Intermediate CNN Layers}
\label{main:layerwise}
Lastly, we examine how downstream performance varies across intermediate representations. Figure~\ref{fig:layer_all} reports task‑averaged $R^2$ and Q‑score as a function of depth for ViT‑Small and ResNet‑50. Negative values are clipped to zero before averaging, and per‑task breakdowns are provided in the appendix.

\textbf{For ViT‑Small}, performance increases over the first few layers and then saturates. Most SSL variants reach near‑peak accuracy by layers 3--5, with only minor changes afterward. Semantic land‑cover targets benefit slightly from deeper layers, whereas continuous biophysical and atmospheric targets (Biomass, Clouds, Heat Island) remain relatively stable across depth and, in some cases (e.g., DINO), degrade modestly in the final layers. This suggests that early transformer layers already capture the information needed for aggregated environmental prediction, while deeper layers increasingly emphasize semantic abstraction.

\textbf{ResNet‑50}, in contrast, exhibits a pronounced inverted‑U pattern. Both $R^2$ and Q‑score peak at intermediate stages (layers 2--4) and decline at the final stage. The drop is primarily driven by the continuous biophysical targets, which improve substantially at intermediate layers, reaching performance competitive with ViT representations, before degrading in the deepest layer. Semantic land‑cover targets continue to improve more steadily with depth. A resized final‑layer reference confirms that this effect is not attributable to embedding dimensionality.

Overall, optimal layer depth is task‑dependent. While final‑layer embeddings are commonly used by default, intermediate layers, especially for CNN backbones, often provide stronger and more stable representations for aggregated biophysical targets. These trends hold consistently across both $R^2$ and Q‑score.

\section{Discussion and Conclusion}
\label{sec:discussion}
Our results show that compact EO embeddings can retain strong downstream utility, but their effectiveness depends on both encoder architecture and how information is preserved under spatial compression. 
Task-level analysis reveals varying difficulty across NeuCo-Bench tasks: semantic land-cover proportion targets are reliably captured, whereas continuous biophysical and atmospheric targets (Biomass, Clouds, Heat Island) are substantially more sensitive to architectural and representational choices. Differences in robustness, captured by the Q-score, can reveal variation even when mean $R^2$ appears saturated.

We employ the \textbf{NeuCo-Bench framework}, which relies on linear probing of 1D embeddings and limits downstream model complexity to isolate representation quality. Some regression targets, particularly continuous biophysical variables, contain inherent label uncertainty that may constrain absolute performance but affects all methods equally.

\textbf{Backbone architecture and representation depth} play a decisive role. Consistent with prior findings on the global modeling capacity of transformer architectures~\cite{raghu_vision_2021, noman_rethinking_2024}, ViTs outperform ResNets when using fixed-size embeddings for aggregated scene-level targets. However, optimal depth is task-dependent: ViT performance typically increases and saturates with depth, whereas ResNet features follow an inverted-U pattern, with intermediate stages substantially outperforming the final layer on continuous biophysical targets. This suggests that defaulting to final-layer embeddings can be suboptimal.

\textbf{Self-supervised objectives and aggregation strategies} shape representation quality. Contrastive objectives (e.g., DINO) favor semantic signals, whereas reconstruction-based objectives (e.g., MAE, FGMAE) better capture continuous variation; no single objective dominates, and complementary strengths emerge. Mean pooling provides a robust aggregation strategy, while min and max pooling often discard informative structure. CLS pooling remains competitive for transformers, but concatenation shows limited complementarity beyond mean pooling alone.

\textbf{Concatenation experiments} demonstrate that meaningful gains arise primarily from diversity across SSL objectives rather than token-aggregation strategies. When baseline task profiles differ, concatenation can consolidate complementary strengths and in some cases also yield per-task improvements beyond both baselines.

\paragraph{Practical implications for EO pipelines.}
Transformer backbones with mean pooling provide a strong default for compact scene-level embeddings. For CNNs, intermediate layers may offer superior representations, suggesting value in exporting multi-layer embeddings. When storage allows modest dimensional increases, combining embeddings pretrained with different SSL objectives can improve robustness without requiring access to raw imagery. Overall, our findings support embedding-centric EO workflows as a scalable alternative to end-to-end fine-tuning: compact, precomputed embeddings retain predictive utility when architectural and representational choices are carefully designed.

\newpage
\vspace{2em}

\noindent\textbf{Acknowledgments} \\

\noindent This research is carried out as part of the Embed2Scale project and is co-funded by the EU Horizon Europe program under Grant Agreement No. 101131841. Additional funding for this project has been provided by the Swiss State Secretariat for Education, Research and Innovation (SERI) and UK Research and Innovation (UKRI).

{
    \small
    \bibliographystyle{ieeenat_fullname}
    \bibliography{references}
}

\clearpage
\setcounter{page}{1}
\maketitlesupplementary
\input{supplement}

\end{document}

%% file: preamble.tex
\usepackage{makecell}
\usepackage{booktabs}



%% file: supplement.tex
\appendix

In the supplementary material, we provide additional results and visualizations that complement the main paper. These analyses cover several experimental axes. 

Section~\ref{app:full_final} reports full performance accuracy results per method, along with additional performance plots for final‑layer GeoFM embeddings. This section complements the main analyses of backbone comparisons (Section~\ref{main:backbone}), SSL objective comparisons (Section~\ref{main:ssl}), and spatial pooling strategies (Section~\ref{main:pooling}).
Section~\ref{sec:app:concat} presents additional visualizations for the concatenation experiments discussed in Section~\ref{main:concat}.
Finally, Section~\ref{app:layerwise} provides per‑task results for intermediate‑layer embeddings, extending the layer‑wise analysis in Section~\ref{main:layerwise}.

\section{Final-Layer Embeddings: Full Per-Task Results}
\label{app:full_final}

\noindent\textbf{Overview.}
This section provides the complete per-task $R^2$ results for all evaluated GeoFM backbones (final-layer embedding), self-supervised objectives, and pooling strategies: Table~\ref{tab:acc_mean} reports mean pooling results for SSL4EO pretrained GeoFMs, and similarly Table~\ref{tab:acc_max} and Table~\ref{tab:acc_min} report max and min pooling, respectively. Methods are sorted by average performance (Avg.) to make consistent cross-task trends explicit and to complement the Q-score–based radar plot shown in the main paper. Additionally, in Table~\ref{tab:acc_terramind_small} all pooling method results for TerraMind-Small are reported.

\begin{figure*}[th]
  \centering
  \includegraphics[width=0.6\textwidth]{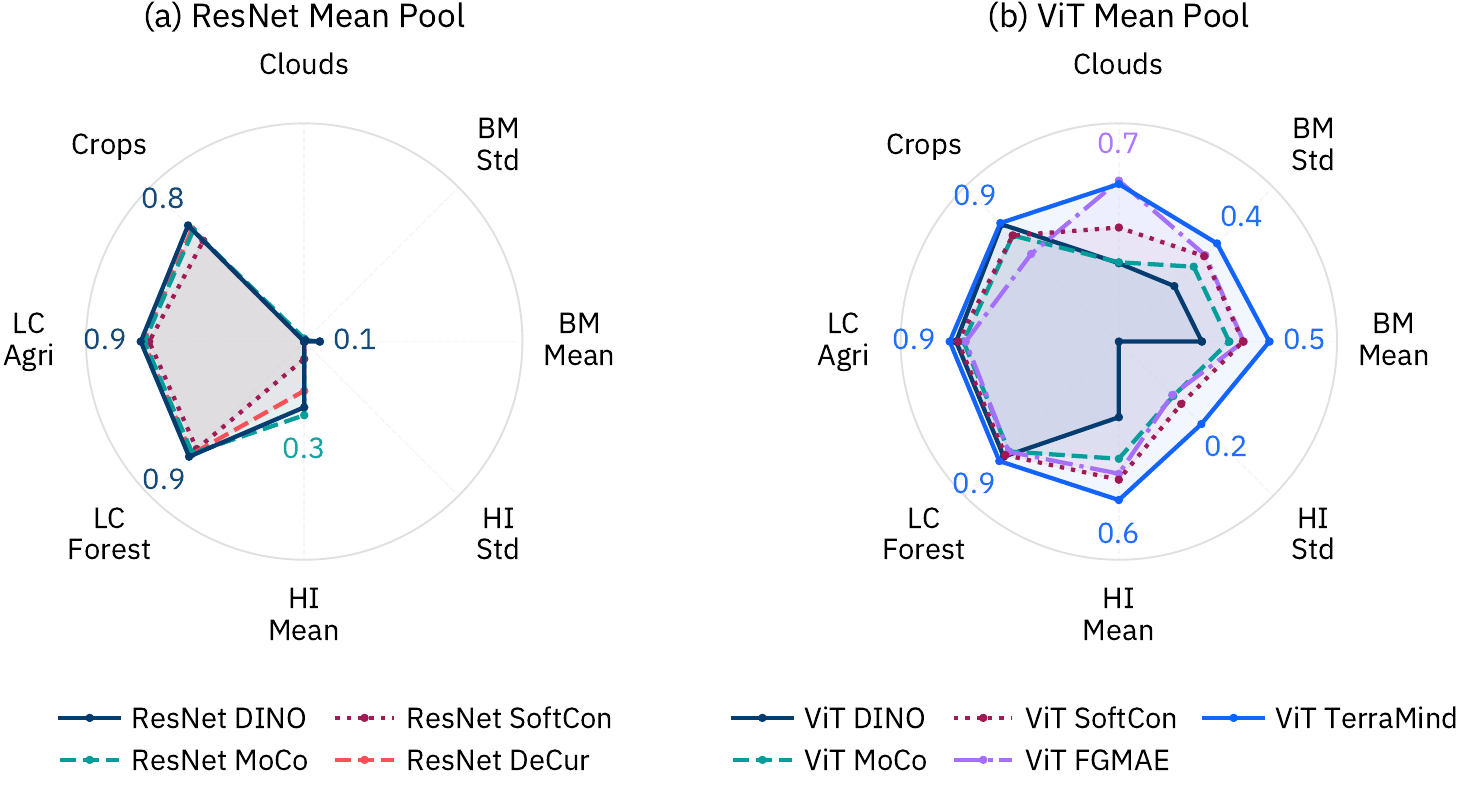}
     \caption{\textbf{Per-task $R^2$ comparison of ResNet-50 (left) and ViT-Small (right) FMs.}
  Final-layer embeddings with mean pooling are used. In contrast to the main paper’s Q-score visualization, this plot reports raw predictive performance ($R^2$) per task. The overall ranking trends remain consistent: ResNet models perform strongly on semantic/land-cover tasks but show limited transfer beyond them, while ViT models are more balanced across tasks. TerraMind remains the most consistent ViT backbone, DINO is particularly strong on land-cover targets, and FGMAE performs well on cloud-cover and biomass tasks. The radial axis is centered at $0$, and the maximum radius is fixed globally with a constant buffer for comparability.}
  
  \label{fig:app:radar_method_r2}
\end{figure*}

\begin{figure*}[th]
  \centering
  \includegraphics[width=0.55\textwidth]{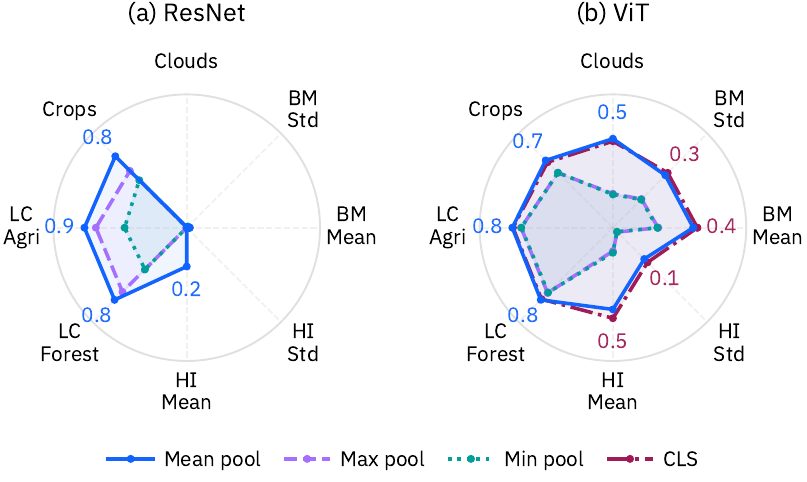}
   \caption{\textbf{Per-task $R^2$ comparison of spatial aggregation methods for ResNet-50 (left) and ViT-Small (right).}
  Final-layer embeddings are evaluated using mean, min, or max pooling (and the CLS token for ViT), with scores averaged across models. This $R^2$ view confirms the Q-score trends reported in the main paper: mean pooling consistently yields the strongest performance across tasks and backbones. For ResNet, max pooling generally outperforms min pooling but both degrade performance relative to mean pooling. For ViT, mean pooling again performs best, with CLS comparable on several semantic tasks, while min and max pooling are similar but systematically weaker—especially on non-land-cover targets. The radial axis is centered at $0$, with a fixed global maximum radius.}
  \label{fig:app:radar_pooling_r2}
\end{figure*}

\noindent\textbf{$R^2$ radar plots.}
Figures~\ref{fig:app:radar_method_r2} and~\ref{fig:app:radar_pooling_r2} replicate the main-paper radar plots using raw $R^2$ instead of Q-score. 
The qualitative trends remain consistent: model families and pooling strategies that rank highly under Q-score also exhibit strong mean predictive performance, confirming that Q-score primarily sharpens separation rather than altering relative rankings. Across both ResNet and ViT backbones, SSL objectives exhibit task-dependent strengths, consistent with the patterns discussed in Section~\ref{main:ssl}. 
No single objective dominates across all tasks, reinforcing the importance of task-aware embedding selection in embedding-centric workflows.

\noindent\textbf{TerraMind results.}
For TerraMind ViT-Small, mean pooling clearly dominates min and max variants, confirming that aggregation choice remains critical even for stronger pretrained backbones.

\noindent\textbf{Pooling comparison.}
Mean pooling consistently yields the strongest performance across backbones. For ViT models, max and min pooling produce similar but consistently lower results. For ResNet models, both max and min pooling substantially degrade performance.

\begin{table*}[t]
  \centering
  \small
  \caption{\textbf{Full per-task $R^2$ scores for tested embedding methods (Mean pooling).} Methods are sorted in ascending order by Avg. For each task, the best-performing method is highlighted in \textbf{bold}, and the second-best is \underline{underlined}.}
  \label{tab:acc_mean}
  \begin{tabular}{lrrrrrrrr|r}
    \toprule
    \textbf{Method}
      & \makecell{\textbf{Biomass}\\\textbf{Mean}}
      & \makecell{\textbf{Biomass}\\\textbf{Std}}
      & \textbf{Crops}
      & \textbf{Clouds}
      & \makecell{\textbf{LC}\\\textbf{Agri}}
      & \makecell{\textbf{LC}\\\textbf{Forest}}
      & \makecell{\textbf{HI}\\\textbf{Mean}}
      & \makecell{\textbf{HI}\\\textbf{Std}}
      & \textbf{Avg.} \\
    \midrule
    ResNet SoftCon (mean) & -0.282 & -0.184 & 0.725 & -0.022 & 0.825 & 0.806 & 0.070 & -0.561 & 0.172 \\
    ResNet DeCur (mean)   & -0.205 & -0.127 & 0.807 & -0.042 & 0.856 & 0.845 & 0.198 & -0.427 & 0.238 \\
    ResNet MoCo (mean)    & -0.139 & -0.125 & 0.798 & 0.013  & 0.851 & 0.838 & 0.296 & -0.332 & 0.275 \\
    ResNet DINO (mean)    & 0.053  & 0.005  & \underline{0.835} & -0.203 & \textbf{0.870} & \textbf{0.863} & 0.264 & -0.282 & 0.301 \\
    ViT DINO (mean)       & 0.282  & 0.217  & \textbf{0.843} & 0.334 & \underline{0.866} & \textbf{0.863} & 0.304 & -0.129 & 0.447 \\
    ViT MoCo (mean)       & 0.375  & 0.293  & 0.762 & 0.338 & 0.827 & 0.824 & 0.471 & \underline{0.158} & 0.506 \\
    ViT MAE (mean)        & 0.408  & \underline{0.335} & 0.609 & \underline{0.684} & 0.800 & 0.804 & 0.530 & 0.145 & 0.539 \\
    ViT FGMAE (mean)      & \textbf{0.424} & \textbf{0.338} & 0.630 & \textbf{0.686} & 0.815 & 0.826 & \underline{0.531} & 0.155 & \underline{0.551} \\
    ViT SoftCon (mean)    & \underline{0.422} & 0.334 & 0.763 & 0.486 & 0.856 & \underline{0.851} & \textbf{0.555} & \textbf{0.181} & \textbf{0.556} \\
    \bottomrule
  \end{tabular}
\end{table*}

\begin{table*}[t]
  \centering
  \small
  \caption{\textbf{Full per-task $R^2$ scores for tested embedding methods (Max pooling).} Methods are sorted in ascending order by Avg. For each task, the best-performing method is highlighted in \textbf{bold}, and the second-best is \underline{underlined}.}
  \label{tab:acc_max}
  \begin{tabular}{lrrrrrrrr|r}
    \toprule
    \textbf{Method}
      & \makecell{\textbf{Biomass}\\\textbf{Mean}}
      & \makecell{\textbf{Biomass}\\\textbf{Std}}
      & \textbf{Crops}
      & \textbf{Clouds}
      & \makecell{\textbf{LC}\\\textbf{Agri}}
      & \makecell{\textbf{LC}\\\textbf{Forest}}
      & \makecell{\textbf{HI}\\\textbf{Mean}}
      & \makecell{\textbf{HI}\\\textbf{Std}}
      & \textbf{Avg.} \\
    \midrule
    ResNet SoftCon (max) & -1.360 & -1.008 & 0.520 & -0.546 & 0.719 & 0.703 & -1.049 & -1.984 & -0.501 \\
    ResNet MoCo (max)    & -0.977 & -0.778 & 0.637 & -0.328 & 0.756 & 0.754 & -0.633 & -1.474 & -0.255 \\
    ResNet DeCur (max)   & -0.781 & -0.674 & 0.677 & -0.386 & 0.776 & 0.770 & -0.641 & -1.637 & -0.237 \\
    ResNet DINO (max)    & -0.723 & -0.589 & 0.683 & -0.498 & 0.770 & 0.758 & -0.519 & -1.336 & -0.182 \\
    ViT MAE (max)        & 0.171  & 0.132  & 0.433 & 0.148  & 0.705 & 0.696 & -0.060 & -0.499 & 0.216 \\
    ViT FGMAE (max)      & 0.155  & 0.110  & 0.451 & 0.101  & 0.709 & 0.711 & -0.070 & -0.390 & 0.222 \\
    ViT DINO (max)       & \textbf{0.368} & \underline{0.240} & \textbf{0.755} & -0.374 & \underline{0.796} & \underline{0.791} & -0.121 & -0.525 & 0.241 \\
    ViT MoCo (max)       & -0.004 & 0.037  & 0.670 & \underline{0.173} & 0.769 & 0.762 & \underline{0.177} & \underline{-0.215} & \underline{0.296} \\
    ViT SoftCon (max)    & \underline{0.321} & \textbf{0.253} & \underline{0.731} & \textbf{0.441} & \textbf{0.836} & \textbf{0.831} & \textbf{0.446} & \textbf{0.074} & \textbf{0.492} \\
    \bottomrule
  \end{tabular}
\end{table*}

\begin{table*}[t]
  \centering
  \small
    \caption{\textbf{Full per-task $R^2$ scores for tested embedding methods (Min pooling).} Methods are sorted in ascending order by Avg. For each task, the best-performing method is highlighted in \textbf{bold}, and the second-best is \underline{underlined}.}
  \label{tab:acc_min}
  \begin{tabular}{lrrrrrrrr|r}
    \toprule
    \textbf{Method}
      & \makecell{\textbf{Biomass}\\\textbf{Mean}}
      & \makecell{\textbf{Biomass}\\\textbf{Std}}
      & \textbf{Crops}
      & \textbf{Clouds}
      & \makecell{\textbf{LC}\\\textbf{Agri}}
      & \makecell{\textbf{LC}\\\textbf{Forest}}
      & \makecell{\textbf{HI}\\\textbf{Mean}}
      & \makecell{\textbf{HI}\\\textbf{Std}}
      & \textbf{Avg.} \\
    \midrule
    ResNet MoCo (min)   & -0.680 & -0.647 & 0.411 & -3.071 & 0.258 & 0.242 & -0.584 & -0.801 & -0.609 \\
    ResNet DeCur (min)  & -1.017 & -0.873 & 0.670 & -2.105 & 0.713 & 0.703 & -0.827 & -1.531 & -0.533 \\
    ResNet DINO (min)   & -1.016 & -0.779 & 0.718 & -1.244 & 0.779 & 0.758 & -0.833 & -1.606 & -0.403 \\
    ResNet SoftCon (min)& 0.043  & 0.026  & 0.304 & -0.253 & 0.317 & 0.238 & -0.020 & \underline{-0.171} & 0.060 \\
    ViT MAE (min)       & 0.131  & 0.107  & 0.460 & 0.121  & 0.694 & 0.673 & -0.018 & -0.456 & 0.214 \\
    ViT FGMAE (min)     & 0.175  & 0.136  & 0.427 & 0.109  & 0.709 & 0.708 & 0.021  & -0.292 & 0.249 \\
    ViT DINO (min)      & \textbf{0.370} & \underline{0.244} & \textbf{0.754} & -0.283 & \underline{0.798} & \underline{0.787} & -0.186 & -0.470 & 0.252 \\
    ViT MoCo (min)      & -0.014 & 0.020  & 0.669 & \underline{0.181} & 0.765 & 0.760 & \underline{0.173} & -0.220 & \underline{0.292} \\
    ViT SoftCon (min)   & \underline{0.324} & \textbf{0.251} & \underline{0.738} & \textbf{0.433} & \textbf{0.840} & \textbf{0.831} & \textbf{0.473} & \textbf{0.067} & \textbf{0.495} \\
    \bottomrule
  \end{tabular}
\end{table*}

\begin{table*}[t]
  \centering
  \small
    \caption{\textbf{Full per-task $R^2$ scores for tested embedding methods (CLS token).} Methods are sorted in ascending order by Avg. For each task, the best-performing method is highlighted in \textbf{bold}, and the second-best is \underline{underlined}.}
  \label{tab:acc_cls}
  \begin{tabular}{lrrrrrrrr|r}
    \toprule
    \textbf{Method}
      & \makecell{\textbf{Biomass}\\\textbf{Mean}}
      & \makecell{\textbf{Biomass}\\\textbf{Std}}
      & \textbf{Crops}
      & \textbf{Clouds}
      & \makecell{\textbf{LC}\\\textbf{Agri}}
      & \makecell{\textbf{LC}\\\textbf{Forest}}
      & \makecell{\textbf{HI}\\\textbf{Mean}}
      & \makecell{\textbf{HI}\\\textbf{Std}}
      & \textbf{Avg.} \\
    \midrule
    ViT DINO (CLS)    & 0.324 & 0.236 & \textbf{0.860} & 0.134 & \textbf{0.878} & \textbf{0.873} & 0.328 & -0.129 & 0.438 \\
    ViT SoftCon (CLS) & 0.369 & 0.286 & 0.769 & 0.404 & \underline{0.855} & 0.850 & 0.504 & 0.105 & 0.518 \\
    ViT MAE (CLS)     & \underline{0.403} & \underline{0.316} & 0.584 & \textbf{0.619} & 0.778 & 0.777 & 0.516 & \underline{0.165} & 0.520 \\
    ViT FGMAE (CLS)   & \textbf{0.413} & \textbf{0.331} & 0.611 & \underline{0.610} & 0.790 & 0.795 & \underline{0.527} & \textbf{0.176} & \underline{0.532} \\
    ViT MoCo (CLS)    & 0.386 & 0.311 & \underline{0.798} & \underline{0.431} & 0.854 & \underline{0.852} & \textbf{0.537} & 0.131 & \textbf{0.537} \\
    \bottomrule
  \end{tabular}
\end{table*}

\begin{table*}[t]
  \centering
  \small
    \caption{\textbf{Full per-task $R^2$ scores for TerraMind ViT-Small (pooling variants).} Results are reported for min, max, and mean pooling and sorted by Avg.}
  \label{tab:acc_terramind_small}
  \begin{tabular}{lrrrrrrrr|r}
    \toprule
    \textbf{Method}
      & \makecell{\textbf{Biomass}\\\textbf{Mean}}
      & \makecell{\textbf{Biomass}\\\textbf{Std}}
      & \textbf{Crops}
      & \textbf{Clouds}
      & \makecell{\textbf{LC}\\\textbf{Agri}}
      & \makecell{\textbf{LC}\\\textbf{Forest}}
      & \makecell{\textbf{HI}\\\textbf{Mean}}
      & \makecell{\textbf{HI}\\\textbf{Std}}
      & \textbf{Avg.} \\
    \midrule
    TerraMind Small (min)  & 0.281 & 0.216 & 0.730 & 0.360 & 0.832 & 0.829 & 0.332 & -0.081 & 0.437 \\
    TerraMind Small (max)  & 0.306 & 0.235 & 0.738 & 0.335 & 0.837 & 0.829 & 0.327 & -0.099 & 0.438 \\
    TerraMind Small (mean) & 0.511 & 0.384 & 0.852 & 0.671 & 0.900 & 0.896 & 0.637 & 0.239 & 0.636 \\
    \bottomrule
  \end{tabular}
\end{table*}

\section{Per-Task Concatenation Results}
\label{sec:app:concat}

This section provides full per-task $R^2$ radar plots for the concatenation experiments introduced in Section~\ref{main:concat}. 
While the main paper reports per-task $\Delta R^2$ bar plots relative to the stronger baseline, here we show the absolute per-task $R^2$ values for both individual embeddings and their concatenation.

\begin{figure*}[th]
  \centering
  \includegraphics[width=\textwidth]{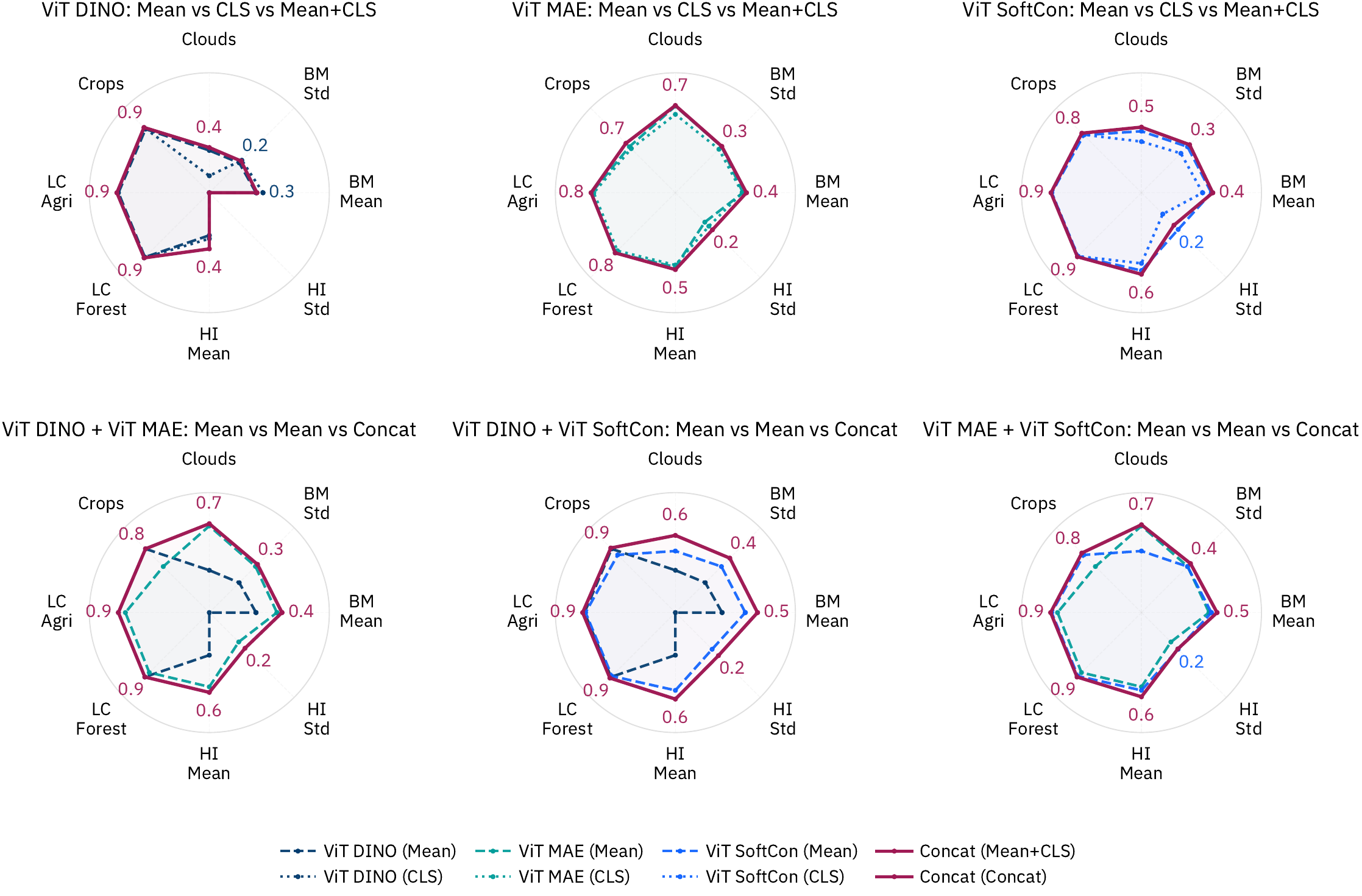}
    \caption{\textbf{Per-task $R^2$ radar plots for embedding concatenation experiments.}
    We report results for all tested combinations, comparing the two individual baselines with their concatenated representation. 
    The plots illustrate that concatenation typically preserves the stronger baseline and yields modest, task-dependent improvements. 
    In particular, combinations such as SoftCon+DINO most consistently show positive deviations over the individual embeddings, suggesting measurable complementarity between pretraining objectives.}
  \label{fig:app:concat}
\end{figure*}

\noindent\textbf{Concatenation analysis.}
As shown in Figure~\ref{fig:app:concat}, concatenation typically preserves the stronger baseline and yields modest, task-dependent gains. In many cases performance remains close to the best single representation, with clearer gains when combining embeddings from different pretraining objectives. Overall, concatenation provides limited but measurable benefits, with selected cases where the joint embedding outperforms the strongest standalone representation on a per-task level.

\section{Per-Task Layerwise Results}
\label{app:layerwise}

\noindent\textbf{Overview.}
This section provides per-task layer-wise performance breakdowns, extending the averaged analysis presented in Section~\ref{main:layerwise}. 
We report both $R^2$ and Q-score to disentangle predictive accuracy and robustness across splits.

\begin{figure*}[th]
  \centering
  \includegraphics[width=\textwidth]{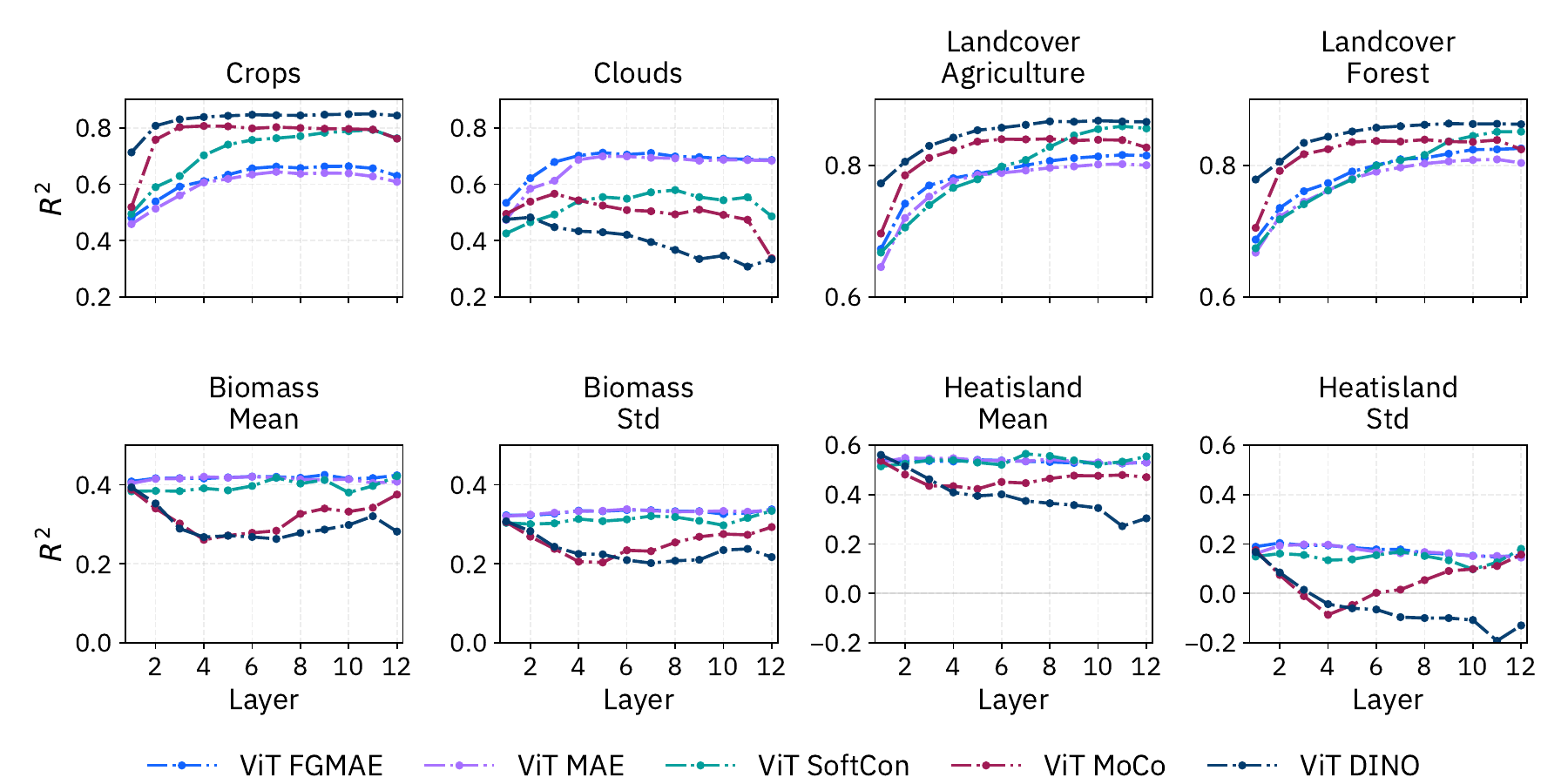}
  \caption{\textbf{Layer-wise per-task downstream performance ($R^2$) for ViT-Small models pretrained on SSL4EO.}
  Results are shown separately per task across layer depth. 
  Semantic and land-cover targets exhibit increasing and saturating trends toward deeper layers, consistent with the averaged analysis in the main paper. 
  Other tasks show early saturation or slight degradation at greater depth.}
  \label{fig:app:layer_vit_r2}
\end{figure*}

\begin{figure*}[th]
  \centering
  \includegraphics[width=\textwidth]{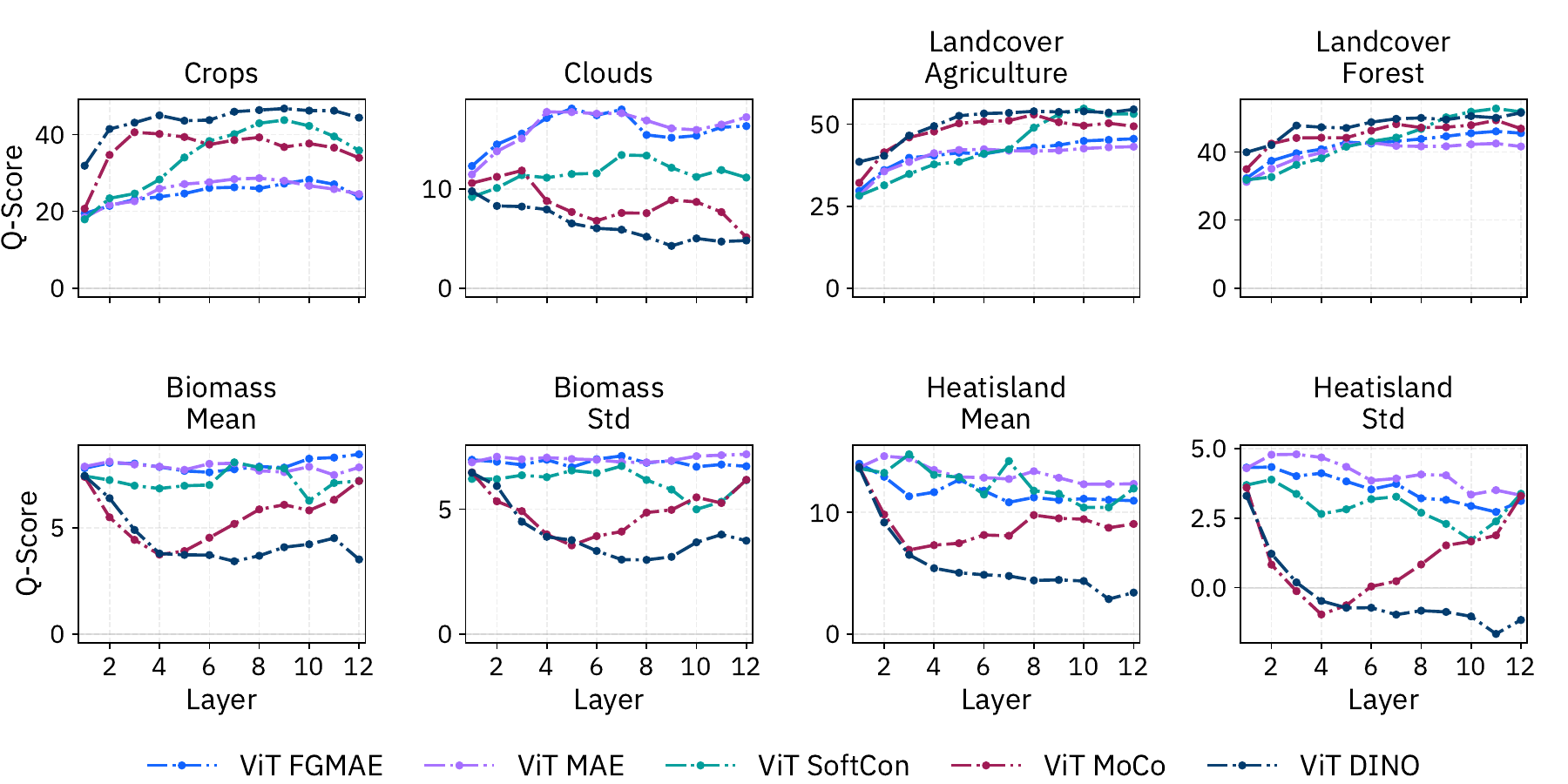}
      \caption{\textbf{Layer-wise per-task downstream performance (Q-score) for ViT-Small models pretrained on SSL4EO.}
      The robustness trends largely mirror the $R^2$ behavior, confirming that depth-dependent effects are consistent across predictive accuracy and stability metrics.}
  \label{fig:app:layer_vit_q}
\end{figure*}

\noindent\textbf{Layer-wise analysis (ViT).}
Figures~\ref{fig:app:layer_vit_r2} and~\ref{fig:app:layer_vit_q} show per-task depth behavior for ViT-Small models. 
Semantic and land-cover tasks exhibit increasing and saturating trends toward deeper layers, mirroring the average performance curves in the main paper. In contrast, several geophysical tasks saturate earlier or show marginal degradation at the deepest layers, indicating that additional depth does not universally improve performance.

\begin{figure*}[th]
  \centering
  \includegraphics[width=\textwidth]{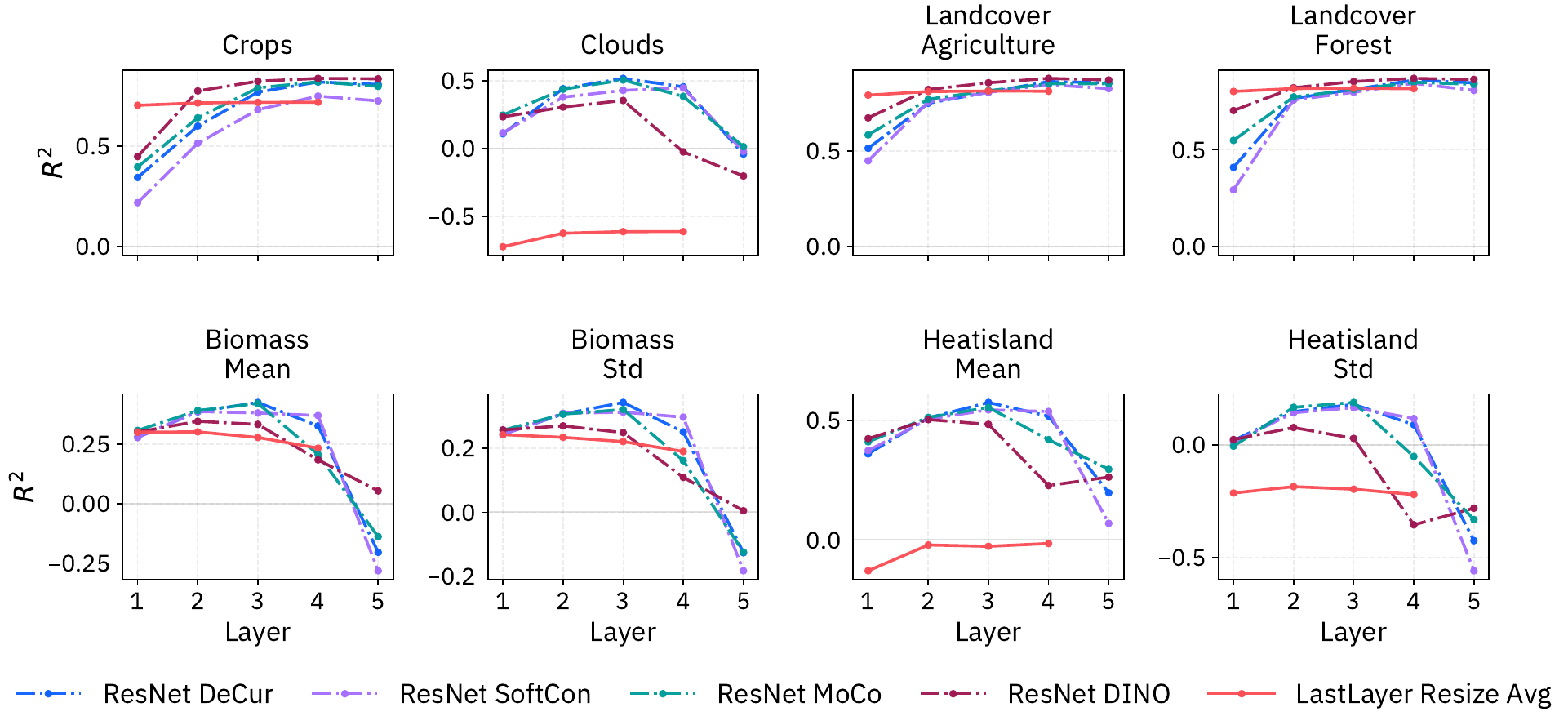}
  \caption{\textbf{Layer-wise per-task downstream performance ($R^2$) for ResNet-50 models pretrained on SSL4EO.}
  Semantic and land-cover tasks show increasing and saturating trends similar to ViT models. 
  In contrast, several other tasks exhibit a pronounced drop at the final-layer, intermediate layers frequently remain competitive with ViT final-layer embeddings.}
  \label{fig:app:layer_resnet_r2}
\end{figure*}

\begin{figure*}[th]
  \centering
  \includegraphics[width=\textwidth]{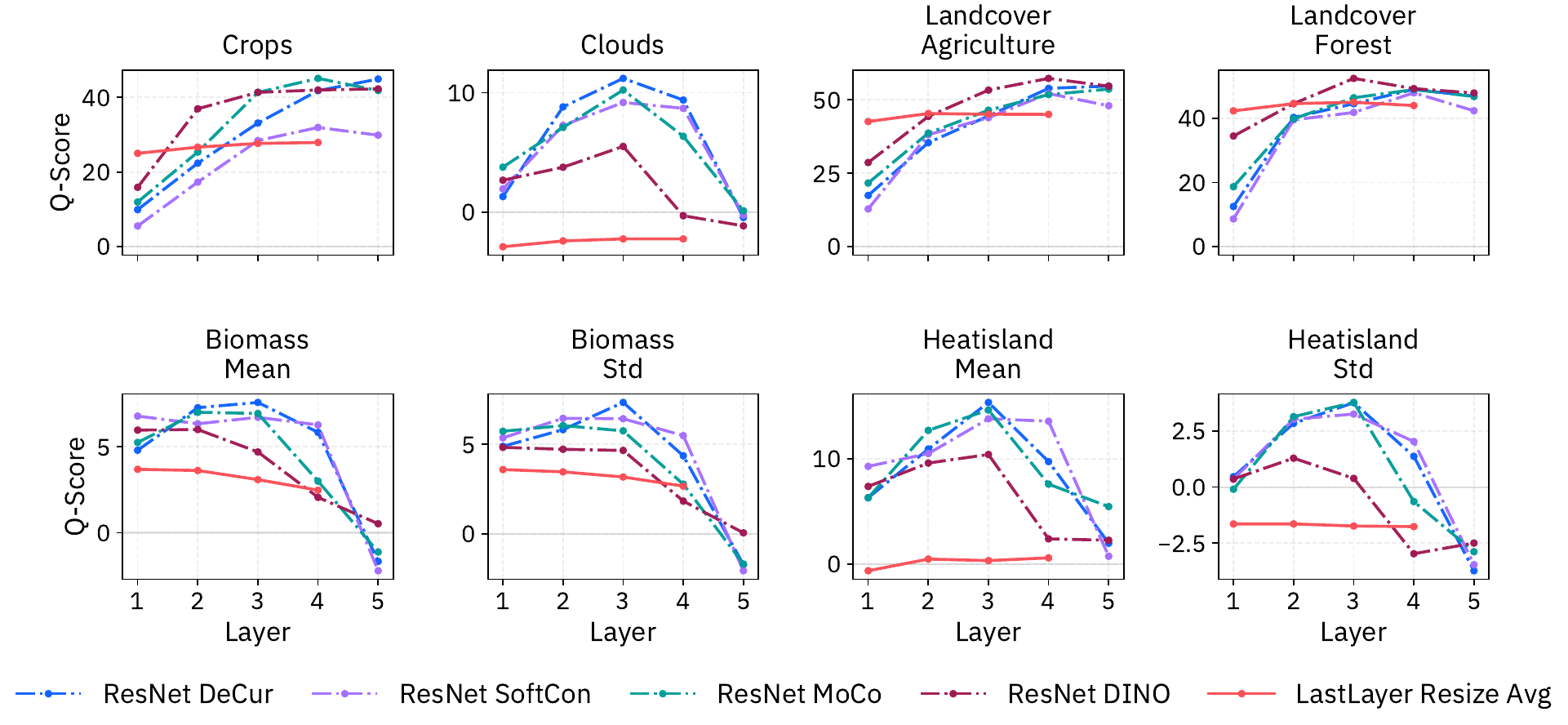}
    \caption{\textbf{Layer-wise per-task downstream performance (Q-score) for ResNet-50 models pretrained on SSL4EO.}
  The robustness metric reinforces the $R^2$ trends, highlighting stronger depth sensitivity in ResNet compared to ViT backbones.}
  \label{fig:app:layer_resnet_q}
\end{figure*}

\noindent\textbf{Layer-wise analysis (ResNet).}
Figures~\ref{fig:app:layer_resnet_r2} and~\ref{fig:app:layer_resnet_q} illustrate a more pronounced depth dependence for ResNet-50 models. While semantic and land-cover tasks benefit from deeper representations, multiple other tasks exhibit a clear performance drop at the final-layer. 
Intermediate layers therefore remain superior for several targets, supporting the main-paper observation that final-layer embeddings can reduce task-agnostic performance for convolutional backbones.

\vspace{2em}

\noindent\textbf{Use of LLMs}\\

\noindent We utilized large language models (LLMs) to refine text and improve readability. All content, including technical material, experimental design, and analyses, was developed by the authors.